\pdfoutput=1

\documentclass[11pt]{article}

\usepackage[preprint]{acl}

\usepackage{latexsym}

\usepackage[T1]{fontenc}
\usepackage{lmodern}  

\usepackage[utf8]{inputenc}

\usepackage{microtype}

\usepackage{inconsolata}
\usepackage{multirow}
\usepackage{subcaption}
\usepackage{comment}

\usepackage{graphicx}   

\usepackage{amsmath,amssymb}
\usepackage{booktabs}   
\usepackage{makecell}

\title{LPI-RIT at LeWiDi-2025: Improving Distributional Predictions via Metadata and Loss Reweighting with DisCo}

\author{
 \textbf{Mandira Sawkar}\thanks{Equal contribution.},
 \textbf{Samay U. Shetty}\footnotemark[1],
 \textbf{Deepak Pandita},
\\
 \textbf{Tharindu Cyril Weerasooriya},
 \textbf{Christopher M. Homan},
\\
\\
 Rochester Institute of Technology
\\
 \small{
   \{ms7201, ss4711, cmhvcs\}@rit.edu,
   \{deepak, cyril\}@mail.rit.edu
 }
}

\begin{document}
\maketitle

\begin{abstract}
The Learning With Disagreements (LeWiDi) 2025 shared task aims to model annotator disagreement through soft label distribution prediction and perspectivist evaluation, which focuses on modeling individual annotators. We adapt DisCo (Distribution from Context), a neural architecture that jointly models item-level and annotator-level label distributions, and present detailed analysis and improvements. In this paper, we extend DisCo by introducing annotator metadata embeddings, enhancing input representations, and multi-objective training losses to capture disagreement patterns better. Through extensive experiments, we demonstrate substantial improvements in both soft and perspectivist evaluation metrics across three datasets. We also conduct in-depth calibration and error analyses that reveal when and why disagreement-aware modeling improves. Our findings show that disagreement can be better captured by conditioning on annotator demographics and by optimizing directly for distributional metrics, yielding consistent improvements across datasets.
\end{abstract}

\section{Introduction}

As machine learning systems increasingly mediate social, legal, and civic decision-making, their alignment with human values becomes paramount. However, as any participant in a democratic process knows well, human disagreement is always present. This includes many existing problems, such as hate speech detection, intent classification, or moral judgment. The LeWiDi 2025 shared task \citep{LeWiDi2025} directly addresses this need by evaluating models on their ability to (1) predict soft label distributions derived from annotator disagreement and (2) approximate individual annotator behavior in a perspectivist setting.

Supervised learning typically resolves annotation disagreement by aggregating labels into a single ground truth, often via plurality vote. However, doing so can obscure valuable minority perspectives, especially on subjective or contentious content \cite{basile-etal-2021-need, prabhakaran-etal-2021-releasing, uma2021learning, plank-2022-problem, Cabitza_Campagner_Basile_2023, homan2023intersectionality, weerasooriya-etal-2023-vicarious, prabhakaran2023framework, pandita-etal-2024-rater}. However, preserving and modeling this disagreement can improve system robustness, fairness, and social accountability. Tasks such as MultiPICo \citep{casola-etal-2024-multipico}, Paraphrase, VariErrNLI, and CSC \citep{jang-frassinelli-2024-generalizable} exemplify domains where capturing nuanced human perspectives, rather than just the majority opinion, is essential for ethical and practical deployment. LeWiDi-2025 challenges systems to go beyond single-label classification and instead model the full distribution of possible human responses.

The core challenge lies in modeling disagreement when annotation is both sparse and noisy. Annotators may vary in reliability, background, and interpretation, and most datasets provide only a few annotations per item. Moreover, models must predict not only soft aggregate distributions but also simulate individual annotator responses, requiring them to generalize from partial supervision over complex, entangled signal sources. Compounding this difficulty is the need for robust evaluation across both soft (e.g., Manhattan, Wasserstein) and perspectivist (e.g., Error Rate, Normalized Absolute Distance) metrics, which test a model’s fidelity to human-like prediction under both collective and individual frames. The four datasets utilized in the shared task are Conversational Sarcasm Corpus (CSC), MultiPico (MP), Paraphrase (Par), and VariErr NLI (Ven).

We adapt the DisCo \citep{weerasooriya-etal-2023-disagreement} model to the LeWiDi 3rd Edition datasets. DisCo consumes item–annotator pairs as input and jointly predicts three interconnected distributions: the specific label an individual annotator would assign, the soft label distribution over all annotators for that item, and the annotator’s own distribution over all items.

While DisCo demonstrated the value of jointly modeling item- and annotator-level distributions, it treated annotators as one-hot IDs and optimized losses misaligned with evaluation. We address both limitations by embedding annotator metadata and by designing loss functions directly tied to disagreement-aware metrics, enabling more interpretable and robust models.

For the post-evaluation phase, we made the following contributions:
\begin{enumerate}
    \item The original DisCo model relied solely on simple annotator ID mappings, limiting its ability to understand annotator characteristics and biases. We modified it to account for annotator metadata features such as age, nationality, gender, education, etc. 
    \item We extended DisCo's preprocessing capabilities to process a wider range of data formats.
    \item We updated the underlying sentence transformer models on which DisCo may depend.
    \item We modified the loss functions to align with the evaluation for soft label distribution prediction and perspectivist modeling.
    \item We perform extensive failure mode analysis on the model.
\end{enumerate}

With these updates, we observed a substantial improvement in the scores for three datasets: CSC, MP, and Par. Additionally, this placed us as rank 4 instead of 7 for Par and Rank 5 instead of 9 for MP in the post-evaluation phase.

\section{Background}

The LeWiDi shared task has emerged as a focal point for advancing methods that embrace, rather than suppress, annotator variation, since its inception \citep{uma-etal-2021-semeval}. The third edition, LeWiDi-2025 \citep{LeWiDi2025}, further extends these efforts by evaluating both distributional and perspectivist modeling across diverse datasets.

LeWiDi-2025 focuses on four core benchmark datasets, each designed to probe different facets of human interpretative variation. Please refer to Appendix~\ref{sec:appendix_datasets} for further information on the datasets.

The LeWiDi evaluation draws on two complementary research traditions. First, item–annotator modeling, the goal is to explicitly account for individual annotator behaviors when aggregating labels. \citet{bd9dfdb6-b296-3318-8bf2-0c827da00fd8}’s foundational model represents each annotator’s reliability via a latent confusion matrix, enabling joint estimation of true item labels and per‐annotator error rates. Subsequent work extended this framework with fully Bayesian treatments \citep{JMLR:v11:raykar10a, pmlr-v22-kim12} and introduced clustering techniques to group annotators by shared labeling patterns \citep{doi:10.1137/1.9781611974010.21}.

In the second paradigm, label distribution learning (LDL) reframes ``ground truth'' not as a single class but as a probability distribution over all possible labels. Under this view, models are trained to match the full annotator‐derived distribution rather than just the majority vote. Early LDL work demonstrated strong performance in tasks like facial age estimation \citep{7439855, Gao_2017} and has since been applied to diverse applications, from short text parsing \citep{shirani-etal-2019-learning} to climate forecasting \citep{doi:10.3233/IDA-184446}, showing that distributional targets can yield richer, more nuanced predictions.

 By learning shared embeddings for both items and annotators, DisCo effectively regularizes sparse annotation settings and pools context across related examples. In experiments on six publicly available datasets, DisCo matched or exceeded state‐of‐the‐art LDL approaches, such as multinomial mixture models combined with CNNs, and outperformed annotator‐modeling baselines like CrowdLayer across both single‐label and distributional evaluation metrics.

Since SemEval-2023, researchers have continued to push toward richer annotator‐aware modeling. IREL \citep{maity-etal-2023-irel} conditions toxicity predictions on anonymized user metadata—integrating each annotator’s identity embedding directly into both the model input and the loss function to improve alignment with individual judgments. CICL\_DMS \citep{grotzinger-etal-2023-cicl}, by contrast, builds on large pre‐trained language models and explores ensemble learning, multi‐task fine‐tuning, and Gaussian process calibration to better match the full distribution of annotator labels. Together, these contributions underscore a growing emphasis on leveraging demographic, behavioral, and contextual signals to capture the nuances of human disagreement.

\section{System Overview}

Our system builds upon the DisCo (Distribution from Context) architecture originally proposed by \citet{weerasooriya-etal-2023-disagreement}. To adapt it for the LeWiDi-2025 task, we
introduced several targeted enhancements, including the use of task-specific sentence encoders, integration of annotator metadata via pretrained embeddings, and modified loss functions to reflect task evaluation metrics. These adaptations enable the model to generalize more effectively from sparse supervision and better capture the complexity of annotator behavior and disagreement. 

DisCo is designed to jointly model individual annotator responses, aggregate item‐level label distributions, and annotator‐level behavior distributions in a unified probabilistic framework.

\begin{figure}[t]
    \centering
    \includegraphics[width=\columnwidth]{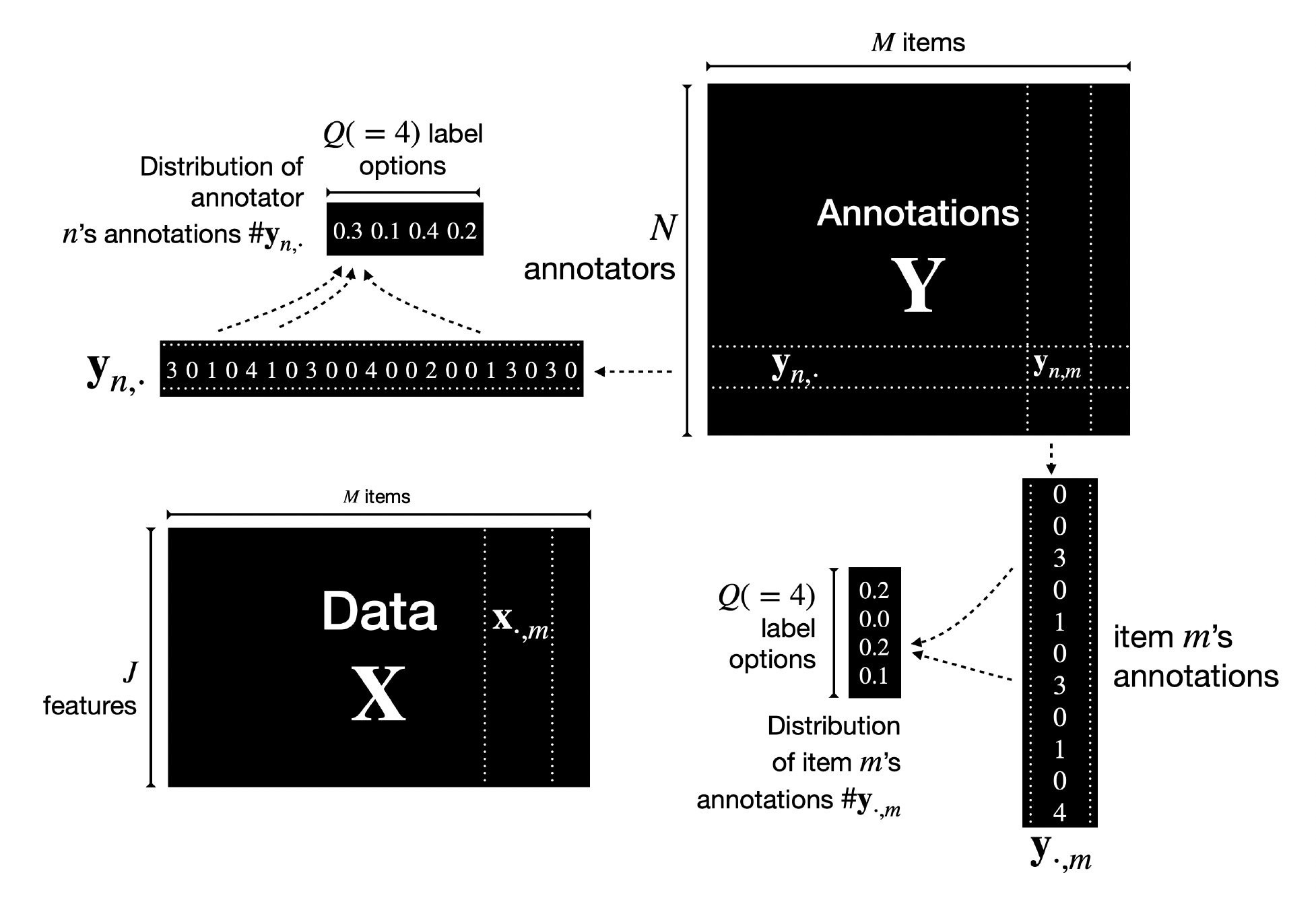}
    \caption{Data representation for DisCo: each item $\mathbf{x}_m$ is paired with per‐annotator responses $\mathbf{y}_{\cdot,m}$ and their empirical distribution $\#\mathbf{y}_{\cdot,m}$, and each annotator $n$ has a response vector $\mathbf{y}_{n,\cdot}$ with distribution $\#\mathbf{y}_{n,\cdot}$.}
    \label{fig:data-representation}
\end{figure}

Each data item $\mathbf{x}_m \in \mathbb{R}^J$ is represented as a column vector of $J$ features, and its associated annotations from $N$ annotators are collected in the matrix $\mathbf{Y} \in \mathbb{Z}_+^{N \times M}$. We denote the vector of responses for item $m$ as $\mathbf{y}_{\cdot,m}$ and the histogram of these responses as $\#\mathbf{y}_{\cdot,m}$. Similarly, each annotator $n$’s behavior across all items is summarized by $\mathbf{y}_{n,\cdot}$ and its histogram $\#\mathbf{y}_{n,\cdot}$. This setup is illustrated in Figure~\ref{fig:data-representation}.

\begin{figure}[t]
    \centering
    \includegraphics[width=\columnwidth]{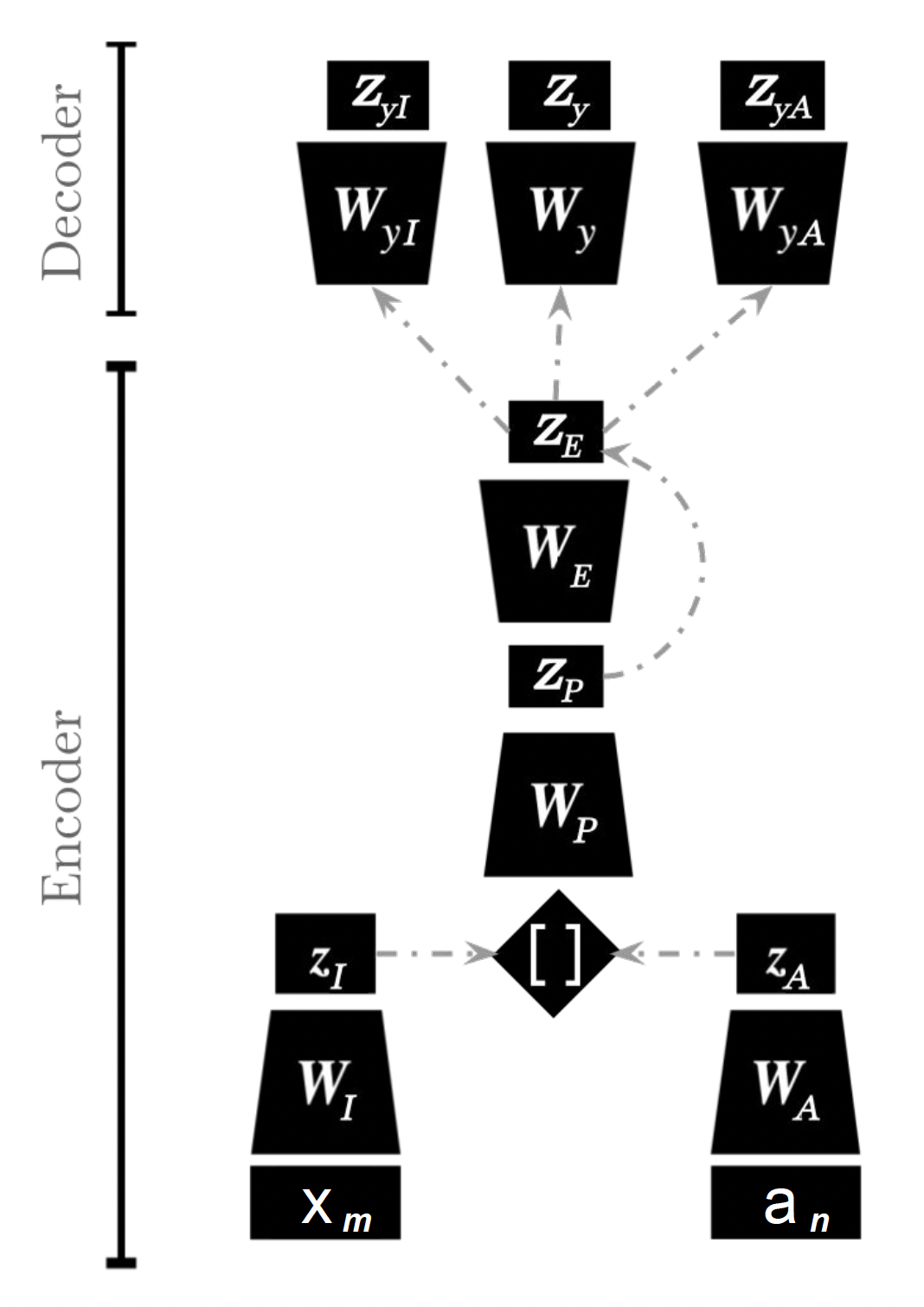}
    \caption{Block diagram of the DisCo encoder and decoder. The encoder maps item and annotator inputs into a joint latent code $\mathbf{z}_E$, and the decoder produces three parallel distributions via softmax heads.}
  \label{fig:block-diagram}
\end{figure}

In the encoder (Figure~\ref{fig:block-diagram}), item and annotator inputs are mapped into separate subspaces. The item vector $\mathbf{x}_m$ is projected via a learnable matrix $\mathbf{W}_I \in \mathbb{R}^{J_I\times J}$ to yield the embedding $\mathbf{z}_I = \mathbf{W}_I \mathbf{x}_m$, while the one‐hot annotator identifier $\mathbf{a}_n$ is projected through $\mathbf{W}_A \in \mathbb{R}^{J_A\times N}$ to produce $\mathbf{z}_A = \mathbf{W}_A \mathbf{a}_n$. These embeddings are concatenated and passed through a two‐layer MLP with softsign activations and a residual connection:
\begin{align}
  \mathbf{z}_P &= \phi\bigl(\mathbf{W}_P\,\cdot \phi([\mathbf{z}_I,\,\mathbf{z}_A])\bigr), \\
  \mathbf{z}_E &= \phi\bigl((\mathbf{W}_E\ \cdot \mathbf{z}_P) + \mathbf{z}_P\bigr),
\end{align}
where $\mathbf{W}_P$ and $\mathbf{W}_E$ are learned projection matrices.

\begin{figure}[t]
    \centering
    \includegraphics[width=\columnwidth]
    {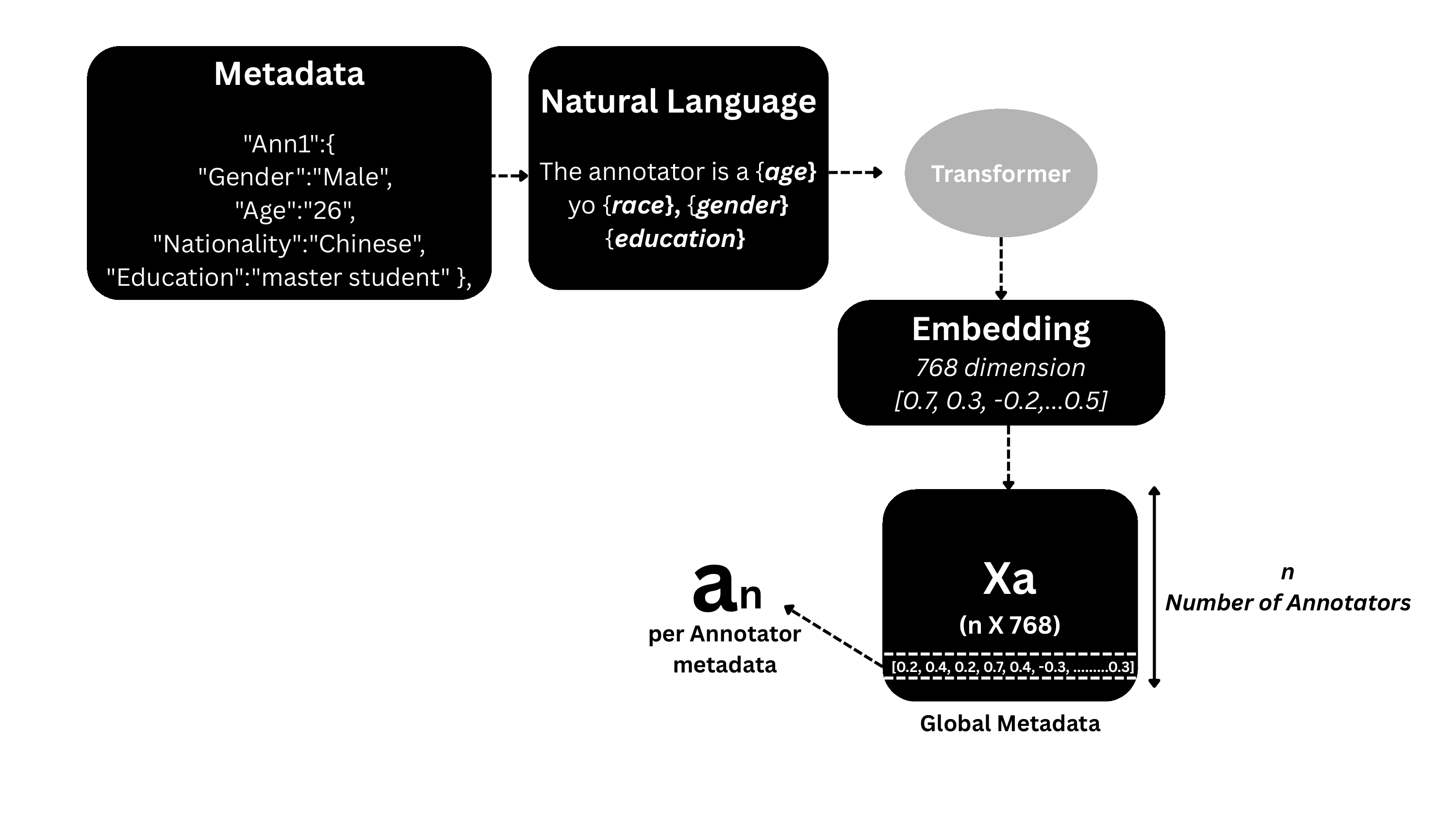}
    \caption{Metadata Embedding Pipeline for DisCo\_New: After converting raw metadata into Natural language, it is passed through a transformer to generate embeddings and eventually generate $a_n$}
    \label{fig:metadata_preprocessing}
\end{figure}

The decoder takes the joint code $\mathbf{z}_E$ and outputs three softmax‐normalized vectors: $\mathbf{z}_y = \mathrm{softmax}(\mathbf{W}_y \mathbf{z}_E)$ for the per‐annotator label distribution $P(y_{n,m}\!\mid\!\mathbf{x}_m,\mathbf{a}_n)$, $\mathbf{z}_{yI} = \mathrm{softmax}(\mathbf{W}_{yI} \mathbf{z}_E)$ for the item‐level distribution, and $\mathbf{z}_{yA} = \mathrm{softmax}(\mathbf{W}_{yA} \mathbf{z}_E)$ for the annotator-level distribution. Training minimizes a composite loss that combines the negative log-likelihood of observed annotator responses with KL divergence terms that align predicted and empirical label distributions at both the item and annotator levels.

At inference time, for an unseen item $\mathbf{x}_m$ without a specific annotator ID, we embed $\mathbf{x}_m$ to obtain $\mathbf{z}_I$ and tile it across all annotator embeddings in $\mathbf{W}_A$ to form $N$ joint codes. Each code is decoded to yield per‐annotator distributions, which are then aggregated by expectation or majority vote to produce the final item‐level prediction. This procedure preserves the learned annotator diversity even when specific annotator metadata is unavailable.

In the post-evaluation phase, we extended the DisCo architecture to better leverage annotator and item information. Annotators were no longer represented by one-hot identifiers but instead by metadata derived from structured JSON inputs. The metadata preprocessing pipeline (Figure~\ref{fig:metadata_preprocessing}) concatenated demographic attributes into a textual description, which was then encoded using a transformer-based sentence embedding model $f_{\text{meta}}(\cdot)$. This produced annotator embeddings $\tilde{\mathbf{a}}_n = f_{\text{meta}}(\text{JSON}(n)) \in \mathbb{R}^D$, which were projected through a learnable matrix $\mathbf{W}_A \in \mathbb{R}^{d_A \times D}$ to yield the annotator representation $\mathbf{z}_A = \mathbf{W}_A \tilde{\mathbf{a}}_n$. On the item side, the generic encoder was replaced with a task-specific transformer encoder $f_{\text{item}}(\cdot)$, producing item vectors $\mathbf{z}_I = \mathbf{W}_I f_{\text{item}}(\mathbf{x}_m)$. Both item and annotator vectors were mapped into semantically aligned subspaces and concatenated into a joint latent representation $\mathbf{z}_E$, which was decoded following the original DisCo framework.

In parallel, we revised the training objective to incorporate additional distributional and per-annotator losses. Beyond categorical negative log-likelihood and KL divergence, we explored Wasserstein distance for soft-label alignment and mean absolute error for per-annotator alignment, as well as combined and alternating formulations. These revisions aligned optimization more closely with the evaluation metrics. Full implementation details, loss formulations, and dataset-level hyperparameter configurations are described in Section~\ref{sec:model_config_new}.

These modifications to the DisCo architecture are not cosmetic but address fundamental gaps: richer annotator modeling and task-aligned optimization.

\section{Experimental Setup}

\subsection{Datasets}

Experiments are conducted on three datasets provided by LeWiDi-2025: Conversational Sarcasm Corpus (CSC), MultiPico (MP), and Paraphrase (Par). Each dataset is provided in a unified JSON format, including item-level features, per-annotator labels, and annotator identifiers. The datasets and their evaluation metrics are discussed further in Appendix~\ref{sec:appendix_datasets}.

\subsection{Tasks}

The system is evaluated on the two complementary tasks defined in the LeWiDi-2025 shared task framework. In \textbf{Task A (Soft Label Prediction)}, a probability distribution over the label space must be output for each instance. Evaluation is conducted using the Manhattan distance for MP and Ven, and the Wasserstein distance for Par and CSC. In \textbf{Task B (Perspectivist Prediction)}, the individual labels assigned by each annotator must be predicted. Evaluation is performed using Error Rate for MP, and Normalized Absolute Distance for Par and CSC. This setup reflects the task’s emphasis on modeling annotator disagreement rather than collapsing it into a single ground-truth label.

\subsection{Model Configuration: DisCo\_OG}

The original DisCo model was adapted to the LeWiDi-2025 tasks with minimal modifications. Annotators were represented using simple identifiers, and the model jointly optimized soft-label and perspectivist objectives. Training used a composite loss combining negative log-likelihood of annotator responses with KL divergence against empirical distributions. Hyperparameters such as activation function, optimizer, dropout rate, learning rate, and fusion strategy were tuned based on validation performance.

\begin{table*}[h]
    \centering
    \small
    \begin{tabular}{lccc}
    \toprule
    \textbf{Hyperparameter} & \textbf{Par Value} & \textbf{MP Value} & \textbf{CSC Value} \\
    \midrule
    Activation & ReLU & Softsign & elu \\
    Annotator Latent Dim & 64 & 64 & 256 \\
    Item Latent Dim & 128 & 256 & 256 \\
    Fusion Type & Concat & Concat & Concat \\
    Optimizer & Adam & Adam & Adam \\
    Learning Rate & 0.001 & 0.001 & 0.001 \\
    Embedding & paraphrase-mpnet-base-v2 & paraphrase-multilingual-mpnet-base-v2 & all-mpnet-base-v2 \\
    Loss & Wasserstein + MAE ($\alpha = 0.6$) & KL Divergence & KL Divergence \\
    Weight Init & Gaussian & Uniform & Gaussian \\
    \bottomrule
    \end{tabular}
    \caption{Best hyperparameters.}
    \label{tab:configs}
\end{table*}

\subsection{Model Configuration: DisCo\_New}
\label{sec:model_config_new}

Building on the architectural extensions described above, we implemented several systematic modifications. 

First, the metadata preprocessing pipeline was redesigned to extract annotator attributes (age, gender, nationality, education, etc.) from structured JSON files. These attributes were verbalized into natural language templates and embedded using transformer-based sentence encoders such as paraphrase-mpnet and all-mpnet. Each annotator’s metadata embedding was 768-dimensional and projected into the model space via a learnable transformation matrix, replacing the simple one-hot identifier scheme used in DisCo\_OG. This richer representation enabled the model to capture systematic annotator behavior beyond identity-level patterns.

Second, the training objectives were expanded. In addition to KL divergence and categorical cross-entropy, we introduced multi-objective loss functions: (i) Wasserstein distance for aligning predicted and true soft-label distributions (applied to Par and CSC), (ii) mean absolute error (MAE) for per-annotator alignment (also on Par and CSC), (iii) a weighted combined loss that optimized both simultaneously, and (iv) an alternating formulation that switched objectives between epochs. The combined loss proved most effective, defined as:
\[
\mathbf{L} = \alpha \cdot \mathbf{L}_{\text{Wasserstein}} + (1 - \alpha) \cdot \mathbf{L}_{\text{MAE}},
\]
with $\alpha = 0.6$ favoring the soft-label component. This formulation produced the most consistent improvements across datasets.

Finally, extensive hyperparameter sweeps were conducted per dataset. The optimal configurations covering activation functions, latent dimensions, fusion strategies, optimizers, learning rates, embedding models, loss functions, and weight initialization schemes are reported in Table~\ref{tab:configs}.

\subsection{Reproducibility}

To ensure reproducibility, all experiments were conducted with fixed random seeds and repeated five times per dataset. The optimal hyperparameter settings for each dataset are reported in Section~\ref{sec:model_config_new}. Source code is publicly available at \url{https://github.com/Homan-Lab/lewidi3_public}. The metadata prompt templates are included in Section~\ref{sec:rep_metadata_prompts} in the appendix to facilitate end-to-end replication of our results.

\section{Results}

We report the official results of our submitted system (under the name ``LPI-RIT'') on the final leaderboard of the LeWiDi 2025 shared task. Table~\ref{tab:results} presents our ranking and evaluation metrics across the three datasets, under both tasks. Our team, ``LPI-RIT'', placed tenth in both soft and perspectivist tasks among fifteen and eleven teams (including LeWiDi baselines), respectively.

\begin{table*}[ht!]
    \centering
    \begin{tabular}{|l|c|c|c|c|c|c|}
    \toprule
    \textbf{Participant} & \multicolumn{3}{c|}{\textbf{TASK A - Soft Evaluation}} & \multicolumn{3}{c|}{\textbf{TASK B - PE Evaluation}} \\
    \cline{2-7}
    & \textbf{CSC} & \textbf{MP} & \textbf{Par} & \textbf{CSC} & \textbf{MP} & \textbf{Par} \\
    \midrule
    taysor & 0.746 & 0.422 & 1.200 & 0.156 & 0.288 & 0.120\\
    dignatev & 0.792 & 0.469 & 1.12 & 0.172 & 0.300 & 0.130 \\
    azadis2 & 0.803 & 0.439 & 1.610 & 0.213 & 0.311 & 0.200 \\
    aadisanghani & 0.803 & 0.439 & 3.050 & 0.213 & 0.311 & 0.490 \\
    twinhter & 0.835 & 0.447 & 0.980 & 0.228 & 0.319 & 0.080 \\
    tomasruiz & 0.928 & 0.466 & 1.800 & 0.231 & 0.414 & 0.230 \\
    \textit{LeWiDi\_mostfrequent} & \textit{1.169} & \textit{0.518} & \textit{3.230} & \textit{0.238} & \textit{0.316} & \textit{0.360} \\
    aadisanghani & 0.803 & 0.439 & 3.051 & 0.213 & 0.311 & 0.491 \\
    funzac & 1.393 & 0.551 & 3.140 & 0.291 & 0.326 & 0.420 \\
    \textbf{LPI-RIT (DisCo\_OG)} & \textbf{1.451} & \textbf{0.540} & \textbf{3.710} & \textbf{0.331} & \textbf{0.324} & \textbf{0.440} \\
    \textit{LeWiDi\_random} & \textit{1.549} & \textit{0.689} & \textit{3.350} & \textit{0.355} & \textit{0.500} & \textit{0.380} \\
    \bottomrule
    \end{tabular}
    \caption{\label{tab:results}
        Final leaderboard scores for LeWiDi 2025. Scores reflect error or distance metrics (lower is better).}
\end{table*}

Compared to the two official baselines, our system outperformed the random baseline across all submitted tasks except for Paraphrase, but performed worse than the most frequent label baseline. For soft labels, our results were 1.45 (CSC), 0.54 (MP), and 3.71 (Par) while in the perspectivist task, they were 0.33 (CSC), 0.32 (MP), and 0.44 (Par).

Despite not achieving top rankings, our system provided a consistent output across tasks and served as a solid implementation of the DisCo modeling framework. These results highlight several areas for improvement—particularly in soft-label prediction on CSC and in modeling individual annotator behavior under the perspectivist setup—while affirming the feasibility of generalizing DisCo to the LeWiDi setting without extensive task-specific modifications.

In the post-evaluation phase, we introduced several improvements to the DisCo model, including the use of annotator metadata, expanded preprocessing support, stronger sentence encoders, and loss functions better aligned with soft-label and perspectivist objectives. These changes led to consistent gains across all datasets. Table~\ref{tab:improved_scores} summarizes these results; further analysis is provided in Section~\ref{sec:discussion}.

\section{Discussion}
\label{sec:discussion}

Having established that DisCo\_NEW consistently outperforms both OG and baselines, we now analyze how and why these improvements occur. In the subsequent comparisons and analyses, the original and updated models are referred to as DisCo\_OG and DisCo\_New, respectively.

\begin{table*}[t]
    \centering
    \small
    \begin{tabular}{lccccc}
    \toprule
    \textbf{Dataset} & \textbf{Task} & \textbf{\_OG Score} & \textbf{\_New Score} & \textbf{LeWiDi Most Frequent Label} & \textbf{LeWiDi Random Label} \\
    \midrule
    \multirow{2}{*}{CSC} & Soft & 1.45 & 0.87 & 1.17 & 1.54 \\
                         & PE & 0.33 & 0.22 & 0.24 & 0.36 \\
    \midrule
    \multirow{2}{*}{MP} & Soft & 0.54 & 0.45 & 0.52 & 0.69 \\
                        & PE & 0.32 & 0.31 & 0.32 & 0.5 \\
    \midrule
    \multirow{2}{*}{Par} & Soft & 3.71 & 2.21 & 3.23 & 3.35 \\
                         & PE & 0.44 & 0.28 & 0.36 & 0.38 \\
    \bottomrule
    \end{tabular}
    \caption{Original vs. new scores across datasets.}
    \label{tab:improved_scores}
\end{table*}

Across all datasets and both tasks, the post-evaluation model (DisCo\_NEW) consistently outperforms both our original submission (DisCo\_OG) and the strongest LeWiDi baselines. On CSC and Par, DisCo\_New reduces error substantially in both soft-label and perspectivist metrics, while on MP the gains are smaller but still clear. These results demonstrate that the proposed extensions—metadata embeddings and task-aligned loss functions—yield tangible improvements over the baseline DisCo architecture and most frequent baselines.

\subsection{MultiPICo Analysis}

\begin{figure}
    \centering
    \includegraphics[width=0.5\textwidth]{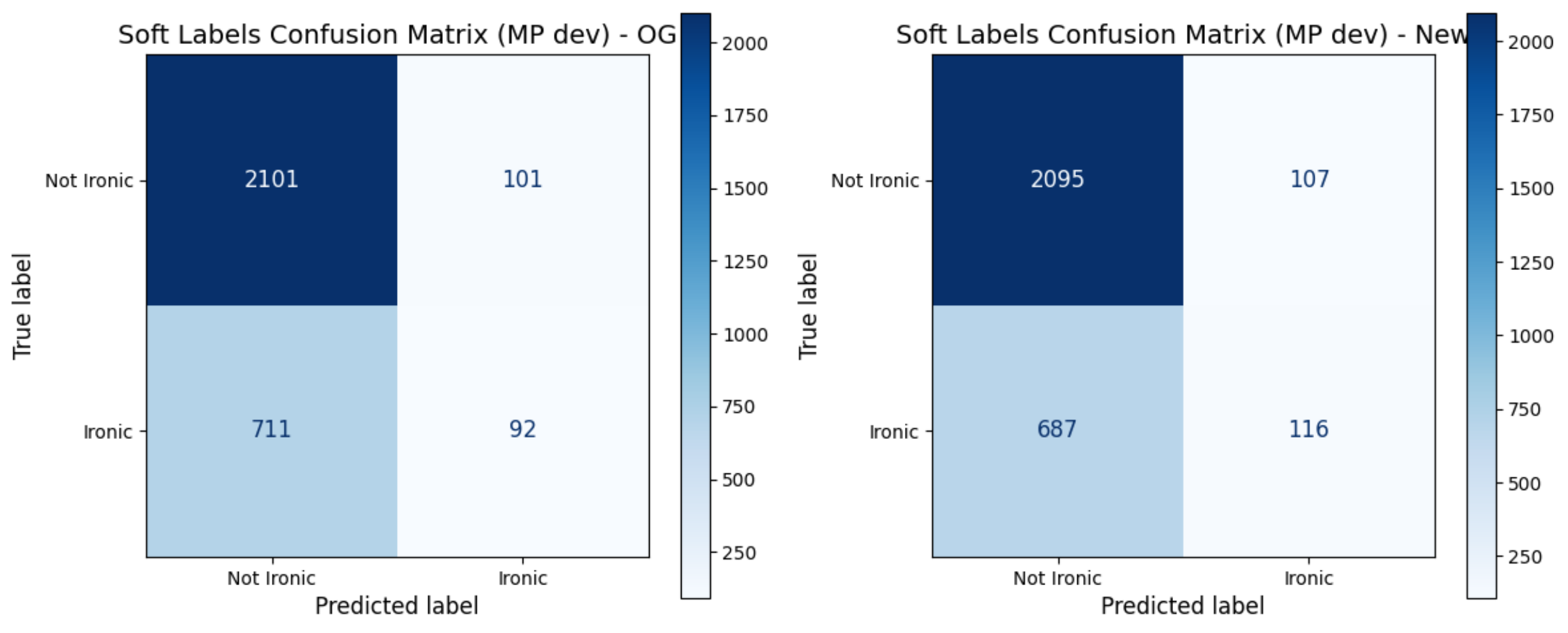}
    \caption{Soft-label confusion matrix for MP dev set (DisCo\_New). Improved recall for the \textsc{Ironic} class is shown compared to DisCo\_OG.}
    \label{fig:mp_conf_matrix}
\end{figure}

\paragraph{Evaluation:}
A modest but consistent reduction in Manhattan distance was observed for DisCo\_New compared to DisCo\_OG (evaluation score reduced from 0.54 to 0.45), indicating that tighter predicted distributions around human soft labels were achieved. A comparison of soft-label confusion matrices (\autoref{fig:mp_conf_matrix}) shows a clear improvement in recall for the \textsc{Ironic} class—true positives increased from 92 to 116, while false negatives decreased from 711 to 687. We interpret this shift as evidence of improved sensitivity to sarcastic and ironic instances, which is a core objective of the MP task. Importantly, these gains were achieved with only a small increase in false positives, suggesting that minority perspectives were captured more effectively without over-predicting irony. The error-rate distribution for individual annotator predictions also improved from 0.32 to 0.31. Overall, stronger alignment at the class level and consistency through replication were demonstrated by DisCo\_New.

\begin{figure}
    \centering
    \includegraphics[width=0.5\textwidth]{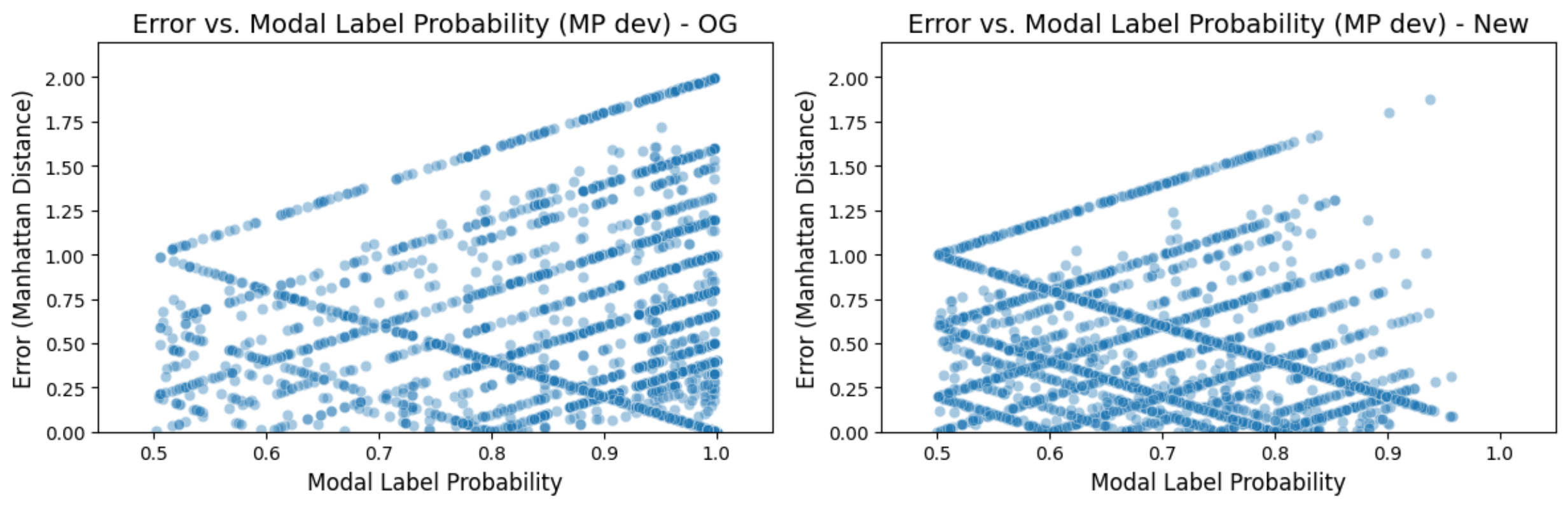}
    \caption{Prediction error vs. modal label probability for the MP dev set. Fewer high-error outliers at high confidence are seen for DisCo\_New.}
    \label{fig:mp_modal_prob_error}
\end{figure}

\paragraph{Confidence Calibration:}
Improvements in model calibration were also observed. Figure (~\ref{fig:mp_modal_prob_error}), a scatterplot of prediction error versus modal label probability, compares model performance DisCo\_OG and DisCo\_New using Manhattan distance against modal prediction confidence. In the original submission, the model exhibited numerous high-error predictions even at high confidence, and the error spread remained large across the confidence spectrum. After improvements, the updated model shows a tighter error distribution, particularly in the 0.7–0.95 confidence range, and fewer catastrophic failures at high confidence. This indicates improved calibration and reliability, although low-confidence predictions continue to produce erratic errors, suggesting room for further refinement in uncertain regions of the prediction space.

\subsection{Paraphrase Analysis}

\paragraph{Evaluation:}
For the Par dataset, the largest improvement in soft-label matching was recorded, with the Wasserstein distance decreasing from 3.71 to 2.21. This indicates substantially better alignment with annotator distributions. The absolute distance was also reduced from 0.44 to 0.28, showing that gains in the soft-label space translated to higher accuracy under the perspectivist evaluation metric. We believe these results demonstrate that DisCo\_New can capture annotator-specific variations more effectively.

\begin{figure}
    \centering
    \includegraphics[width=0.5\textwidth]{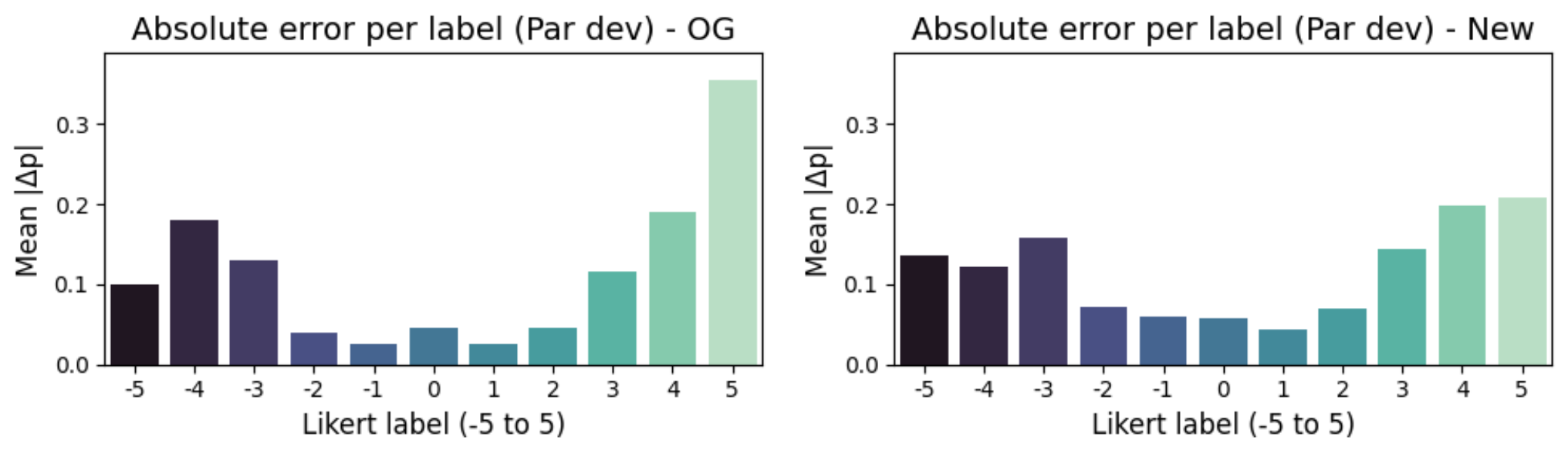}
    \caption{Mean absolute error per Likert label on the Par dev set. DisCo\_New (blue) shows a more balanced and lower error profile, especially at the extremes.}
    \label{fig:par_label_error}
\end{figure}

\paragraph{Error Calibration by Label:}
To assess model behavior across the Likert scale, mean absolute error per label was examined. As shown in Figure~\ref{fig:par_label_error}, predictions from DisCo\_OG were highly skewed, with excessive probability mass assigned to label \texttt{+5}, producing sharp error peaks. A more balanced error profile was seen in DisCo\_New, with reduced overcommitment to extreme positive labels while calibration error in the mid-range was maintained or slightly increased. This suggests that output bias was corrected in a way that more faithfully reflects the true distribution of paraphrase strength.

\begin{figure}
    \centering
    \includegraphics[width=0.5\textwidth]{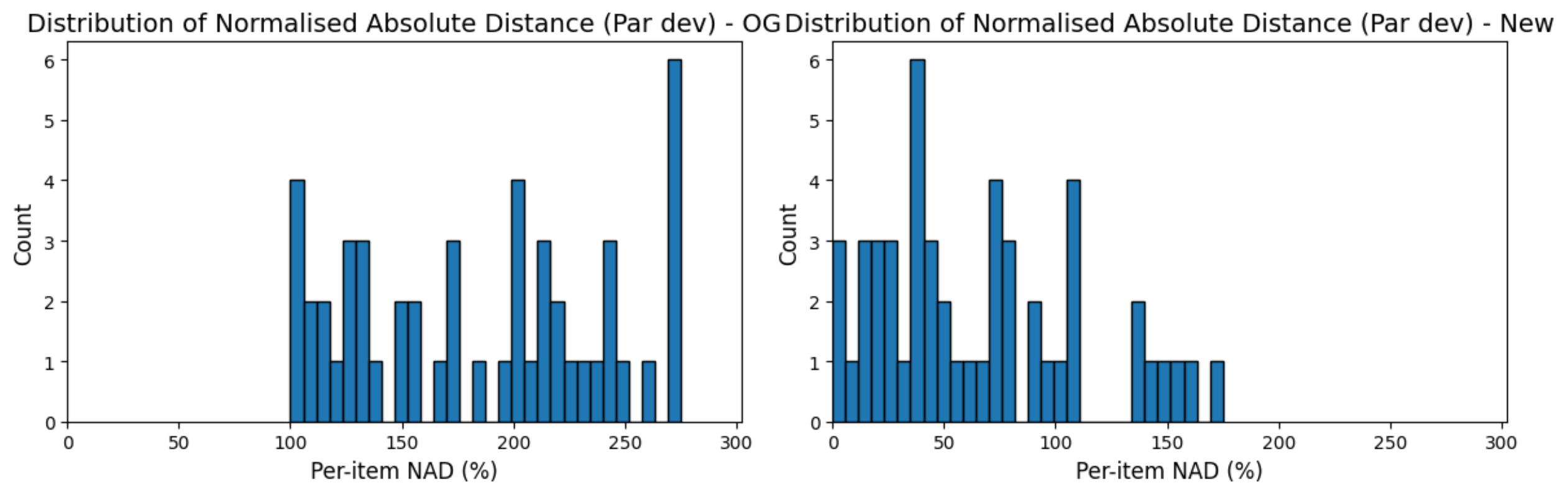}
    \caption{Distribution of Normalized Absolute Distance (NAD) for the Par dev set. DisCo\_New exhibits a sharper peak and lower error across the board.}
    \label{fig:par_nad}
\end{figure}

\paragraph{Normalized Error Distribution:}
Overall soft-label alignment was further assessed using Normalized Absolute Distance (NAD), which measures deviation from the gold distribution relative to total mass. As shown in Figure~\ref{fig:par_nad}, lower and more concentrated NAD scores were achieved by DisCo\_New, with most predictions deviating less than 75\%. In contrast, DisCo\_OG exhibited inflated NAD values due to label scale mismatches and miscalibration. We view this as evidence that DisCo\_New better captures the inherent ambiguity and subjectivity in paraphrase judgments.

\subsection{Conversational Sarcasm Corpus (CSC)}

\paragraph{Evaluation:}
For CSC, clear gains in soft-label alignment were recorded. The Wasserstein distance decreased from 1.45 in DisCo\_OG to 0.87 in DisCo\_New, indicating a closer approximation to gold label distributions. This improvement was especially evident for examples with low annotator consensus. The absolute distance also fell from 0.33 to 0.22, showing significant enhancement in the perspectivist task.

\begin{figure}
    \centering
    \includegraphics[width=0.5\textwidth]{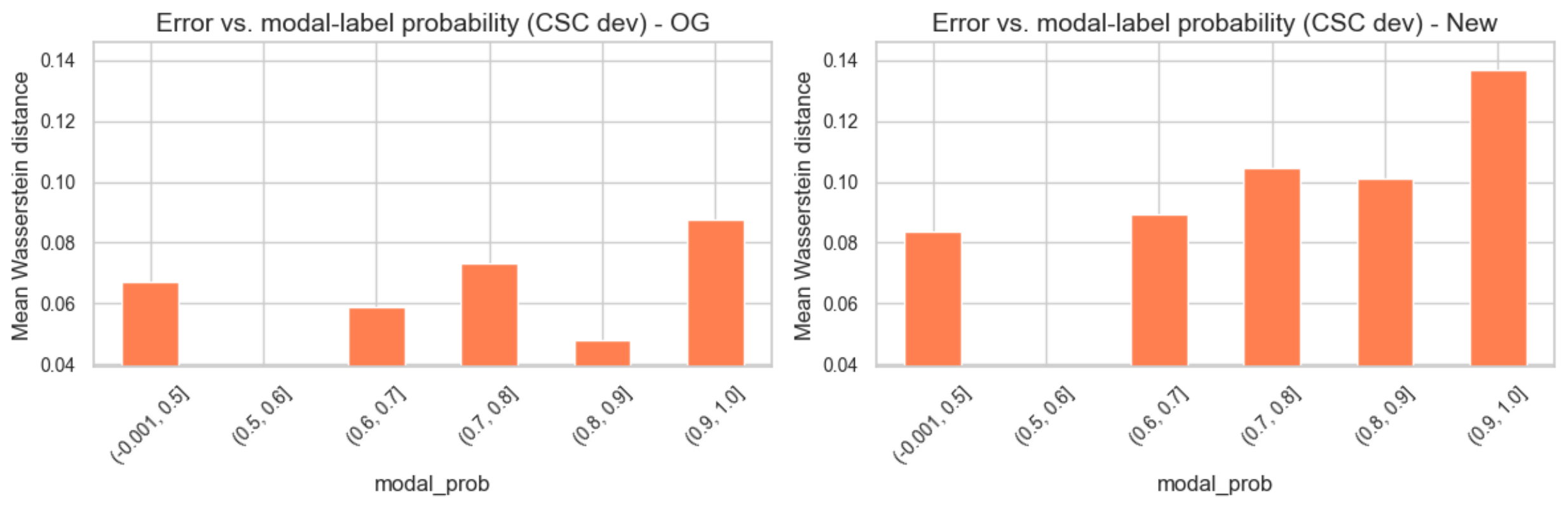}
    \caption{Prediction error vs. modal label probability on the CSC dev set. Reduced error on low-agreement cases is observed for DisCo\_New.}
    \label{fig:csc_modal_prob_error}
\end{figure}

\paragraph{Confidence Sensitivity:}
The effect of gold label certainty on model performance was examined by plotting prediction error against modal label probability. As shown in Figure~\ref{fig:csc_modal_prob_error}, lower error for cases with low modal confidence (high annotator disagreement) was achieved by DisCo\_New. While DisCo\_OG exhibited the highest Wasserstein error in these ambiguous cases, DisCo\_New maintained greater stability and resilience, capturing soft-label nuances even when consensus was weak. We see this as further support for the model’s improved perspectivist capabilities and robustness in handling disagreement.

\begin{figure}
    \centering
    \includegraphics[width=0.5\textwidth]{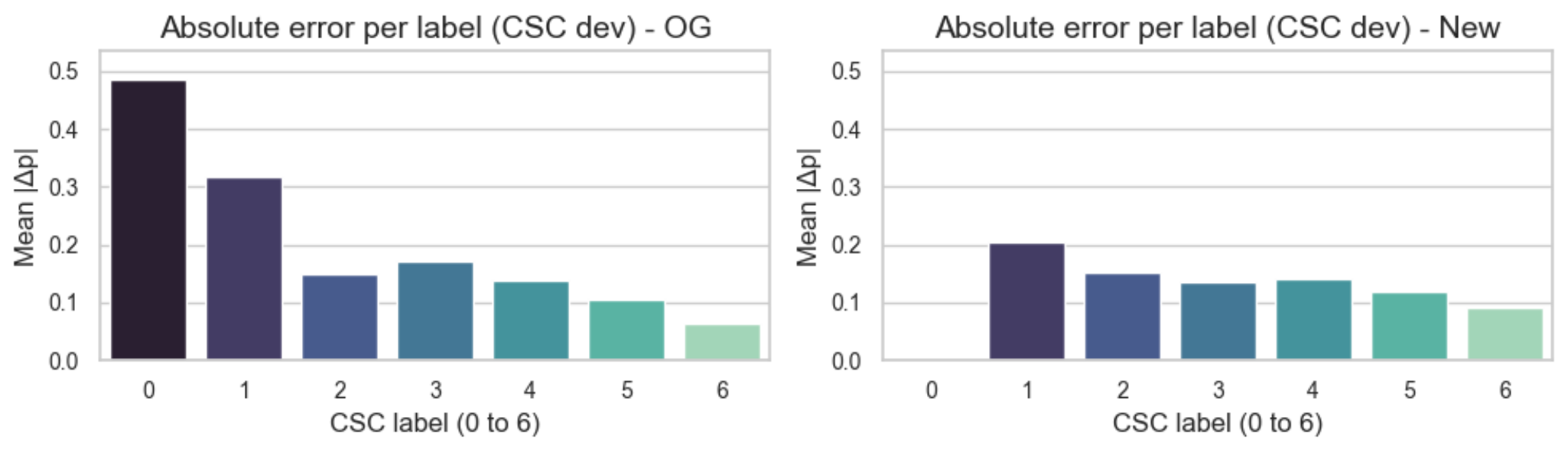}
    \caption{Mean absolute error per Likert label on the CSC dev set. DisCo\_New reduces overprediction of non-sarcastic responses (label 0) and achieves smoother calibration overall.}
    \label{fig:csc_label_error}
\end{figure}

\paragraph{Error Calibration by Label:}
Mean absolute error per Likert label (Figure~\ref{fig:csc_label_error}) showed that DisCo\_OG over-predicted label \texttt{0}—non-sarcastic interpretations—resulting in large mismatches. This overcommitment was reduced by more than half in DisCo\_New. A smoother error profile across all sarcasm intensities was also observed, avoiding the sharp asymmetries seen in DisCo\_OG. These findings indicate a more balanced and context-aware handling of literal and sarcastic language, with improved soft-label calibration overall.

\subsection{Cross-Dataset Insights}

Several cross-cutting patterns emerged across CSC, MP, and Par, providing broader insight into the handling of label ambiguity, annotator disagreement, and error sensitivity.

\begin{figure}[!htb]
    \centering
    \begin{subfigure}{0.7\linewidth}
        \centering
        \includegraphics[width=\textwidth]{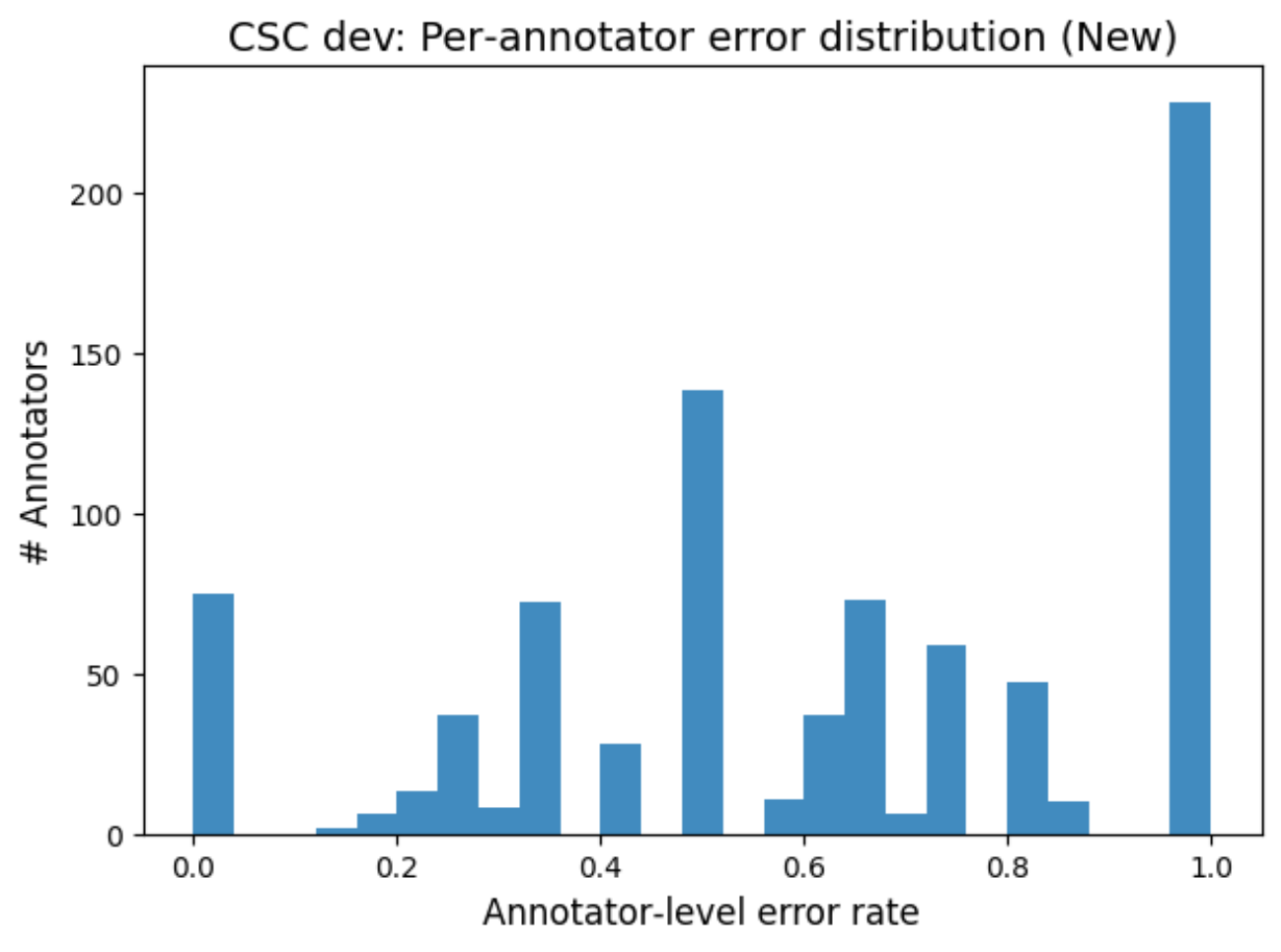}
        \caption{CSC (New)}
    \end{subfigure}
    \begin{subfigure}{0.7\linewidth}
        \centering
        \includegraphics[width=\textwidth]{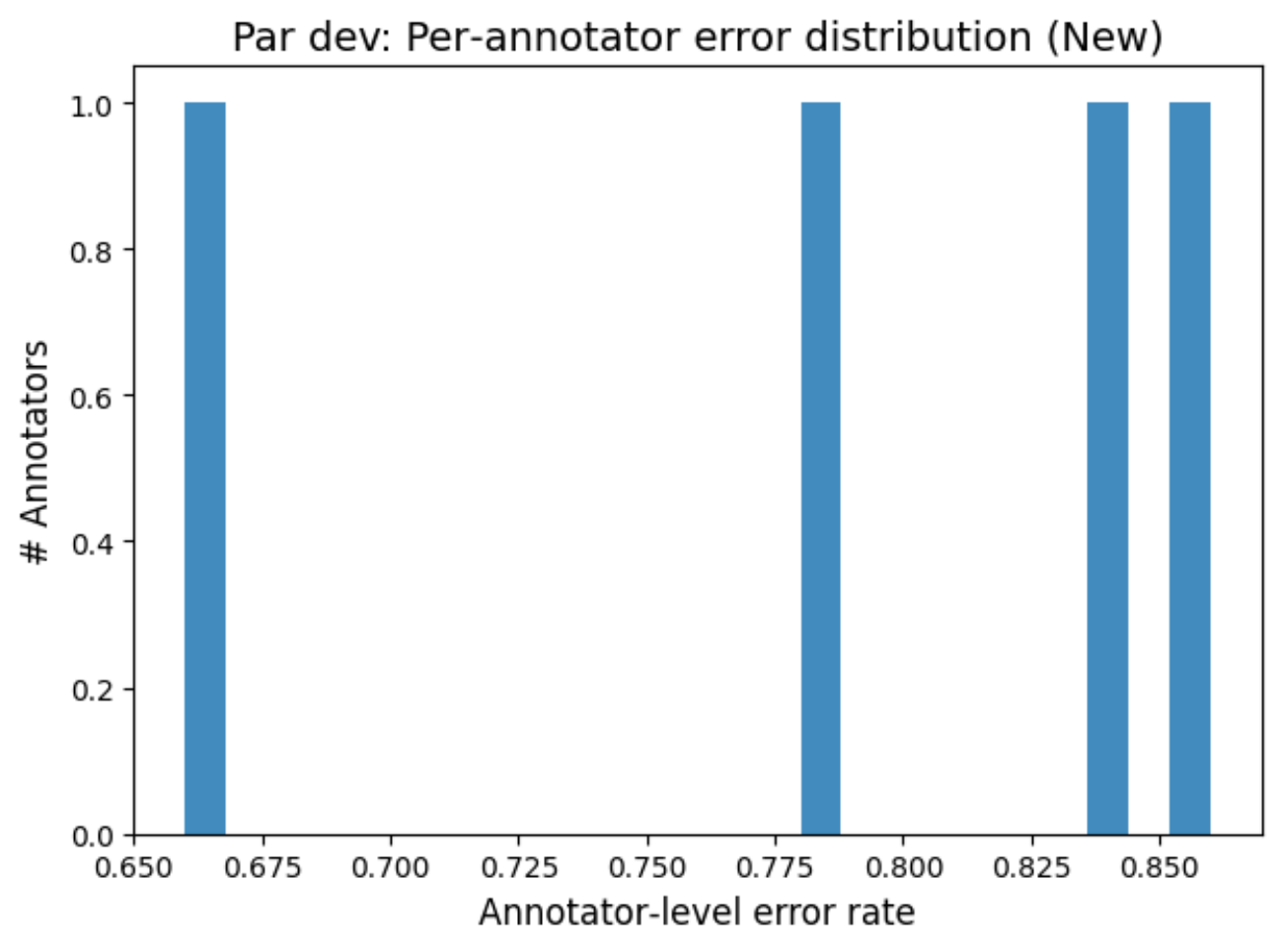}
        \caption{MP (New)}
    \end{subfigure}
    \begin{subfigure}{0.7\linewidth}
        \centering
        \includegraphics[width=\textwidth]{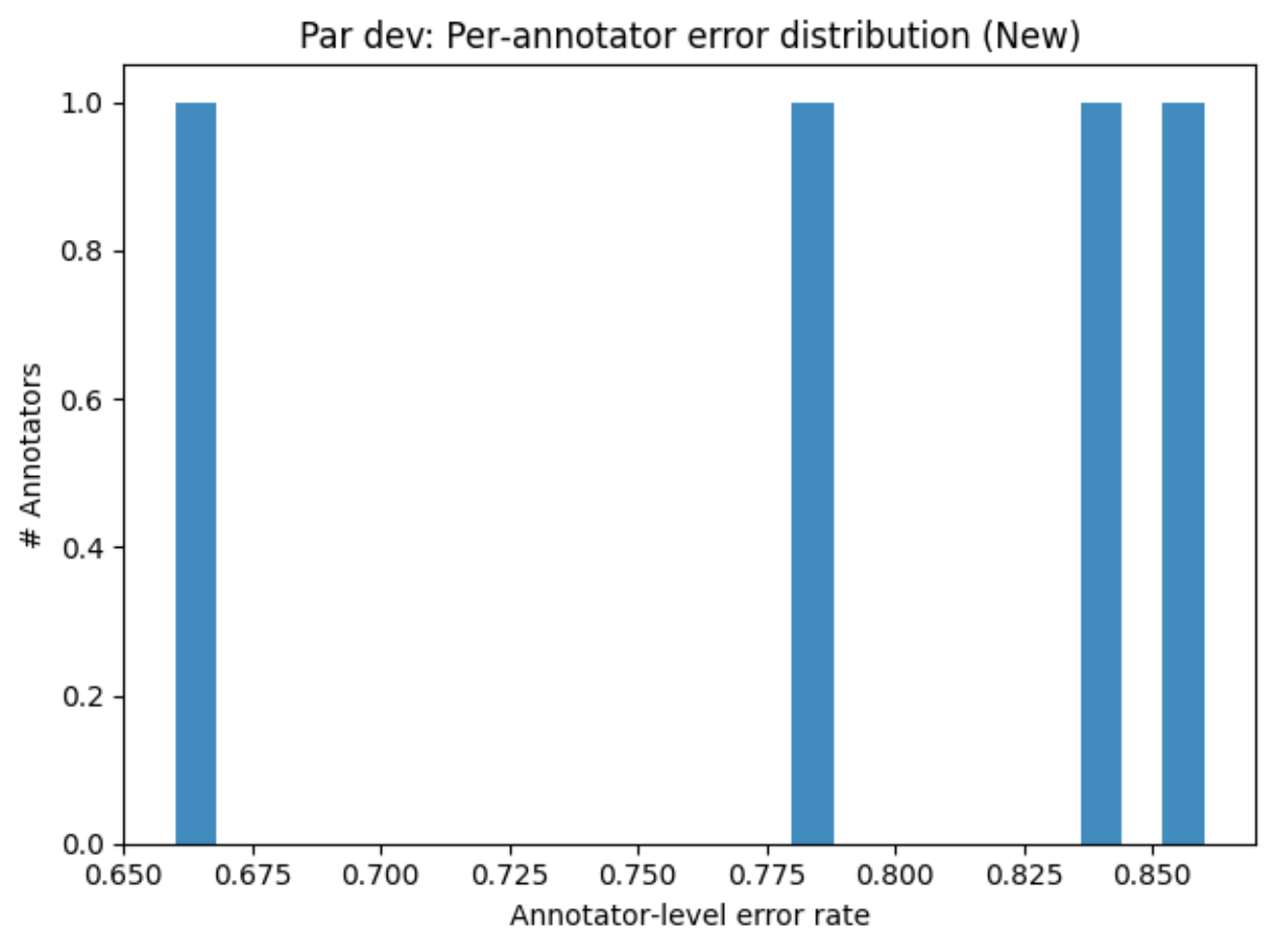}
        \caption{Par (New)}
    \end{subfigure}
    \caption{Annotator-level error distributions for the DisCo\_New model. Each histogram shows the distribution of absolute error per annotator across the dataset.}
    \label{fig:annotator_error_dists_split}
\end{figure}

\paragraph{Annotator-Level Evaluation:}
Annotator error distributions (Figure~\ref{fig:annotator_error_dists_split}) showed that for CSC, virtually all annotators were predicted incorrectly by DisCo\_OG—error rates clustered at 1.0. In contrast, a more varied distribution was seen for DisCo\_New, with many annotators achieving error rates below 0.6. We interpret this as evidence of better alignment with annotator-specific viewpoints. MP remained largely stable, with a slightly tighter distribution under DisCo\_New. For Par, high error persisted in both models, driven by strong prior bias in predictions. These findings confirm that while overall system-level scores improved modestly, substantial gains in modeling annotator diversity and disagreement were achieved for CSC.

Additional linguistic and entropy-based analyses in Appendix~\ref{sec:appendix_supplementary_analysis} further support these findings.

These analyses show that beyond leaderboard scores, disagreement-aware modeling yields interpretable and socially relevant gains.

\section{Conclusion}

We presented enhancements to the DisCo architecture in the context of the LeWiDi-2025 shared task, addressing key limitations in annotator modeling, input representation, and loss formulation. By embedding annotator metadata, refining item encoders, and introducing task-aligned multi-objective losses, our post-evaluation system achieved consistent improvements across CSC, MP, and Par in both soft-label and perspectivist evaluations.

Beyond leaderboard performance, our analyses revealed important behavioral patterns: improved calibration under uncertainty, stronger alignment with annotator-specific perspectives, and greater robustness to label ambiguity. These findings demonstrate that modeling disagreement is not only a technical challenge but also an opportunity to capture the diversity inherent in human annotation.

Looking ahead, we see promising directions in scaling demographic-aware modeling, developing systematic ablation studies, and exploring methods that safeguard fairness and privacy while leveraging annotator metadata. Our work underscores the value of moving beyond aggregated ground truth toward systems that better reflect the complexity of human judgment.

\section*{Limitations}

Our study has some limitations. First, we did not evaluate on the VariErrNLI dataset, primarily due to time constraints and the additional modeling adjustments the dataset features would require. As a result, our findings are restricted to CSC, MP, and Par, and may not fully generalize to NLI-style disagreement tasks.

Second, while our system integrates multiple extensions to DisCo, including metadata embeddings and revised loss formulations, we did not conduct full ablation studies. Consequently, it is difficult to isolate the contribution of each component, and future work should aim to quantify their relative impact more systematically.

Finally, the use of annotator metadata raises ethical considerations. Demographic information such as age, gender, and nationality can be valuable for modeling disagreement, but also introduces potential risks around privacy and fairness if applied in real-world systems. These aspects warrant further investigation before deployment in sensitive applications.

Future work should address these limitations by extending evaluation to broader datasets, performing systematic ablations, and developing methods that leverage annotator metadata while safeguarding privacy and fairness.

\section*{Acknowledgments}

We thank the organizers of the LeWiDi-2025 shared task for providing the datasets, evaluation framework, and leaderboard infrastructure. We are also grateful to the anonymous reviewers for their constructive feedback, which helped improve this work. Additionally, we thank Ayo Owolabi for the insightful discussions.

\nocite{Ando2005,andrew2007scalable,rasooli-tetrault-2015}

\bibliography{anthology, custom, custom_dp}

\appendix

\section{Appendix}

\subsection{Datasets}
\label{sec:appendix_datasets}

\paragraph{Conversational Sarcasm Corpus (CSC):} It comprises roughly 7,000 context–response pairs, each annotated for sarcasm intensity on a six‐point scale by both the original response generators (“speakers”) and subsequent external observers \citep{jang-frassinelli-2024-generalizable}. In an initial online experiment, speakers wrote a reply to a given situational context and self‐rated the sarcasm of their own utterance from 1 (“not at all”) to 6 (“completely”). In follow-up studies, fresh cohorts of observers provided independent ratings for the same context–response pairs—six observers per item in Part 1 and four in Part 2—yielding rich soft label distributions that reflect both insider and outsider perspectives.

\paragraph{MultiPico (MP):} The dataset is a multilingual irony‐detection corpus built from short post–reply exchanges drawn from Twitter and Reddit \citep{casola-etal-2024-multipico}. For each entry, crowdsourced annotators judged whether the reply was ironic in light of the preceding post, producing a binary label. Crucially, MP includes sociodemographic metadata (gender, age, nationality, race, student/employment status) for each annotator, and covers eleven languages—among them Arabic, Dutch, English, French, German, Hindi, Italian, Portuguese, and Spanish. On average, each post–reply pair receives five independent annotations, making MP a challenging benchmark for cross‐lingual and demographic‐aware perspectivist modeling. The paper describing this dataset is available here.

\paragraph{Paraphrase Detection (Par):} The benchmark adapts the Quora Question Pairs (QQP) format to a fine‐grained judgment task. Four expert annotators each assigned an integer score from –5 (“completely different”) to +5 (“exact paraphrase”) for 500 question pairs, and provided brief justifications for their ratings. Unlike typical NLI‐style datasets, Par uses scalar labels and limits each annotator to one judgment per item, emphasizing inter‐annotator variance in graded semantic similarity. This dataset is maintained by the MaiNLP Lab and is not yet formally published.

\paragraph{VariErr NLI ((VariErrNLI)):} The corpus was specifically designed to disentangle genuine human label variation from annotation errors in Natural Language Inference (NLI) tasks \citep{weber-genzel-etal-2024-varierr}. In the first round, annotators re‐labeled 500 premise–hypothesis pairs drawn from the MNLI corpus, providing both labels (Entailment, Neutral, or Contradiction) and free‐text explanations for their choices. In the second round, these same annotators validated each label–explanation pair, yielding 7,732 judgments that pinpoint error versus variation. LeWiDi-2025 focuses on the Round 1 soft label distributions, challenging systems to model nuanced NLI judgments at the intersection of semantics and annotator reasoning. The paper describing this dataset is available here.

\subsection{Supplementary Analysis}
\label{sec:appendix_supplementary_analysis}

This section provides additional analyses for the three datasets, supplementing the main results discussed in Section~\ref{sec:discussion}. The figures below explore linguistic complexity, annotator alignment, and perspective variance in greater detail.

\subsubsection{Qualitative Insights from Word Clouds:}
\label{sec:wordclouds}

\begin{figure}[!htb]
    \centering
    \begin{subfigure}{0.9\linewidth}
        \centering
        \includegraphics[width=\textwidth]{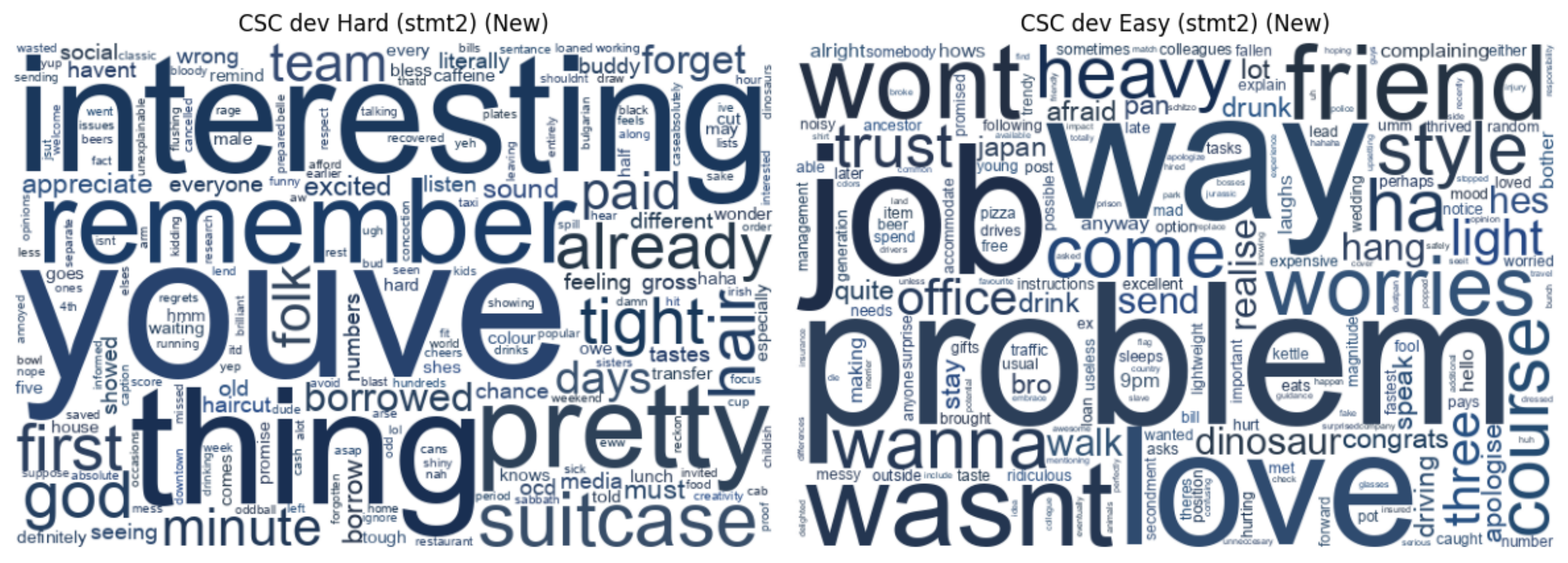}
        \caption{CSC (New)}
    \label{fig:wordcloud_csc}
    \end{subfigure}
    \begin{subfigure}{0.9\linewidth}
        \centering
        \includegraphics[width=\textwidth]{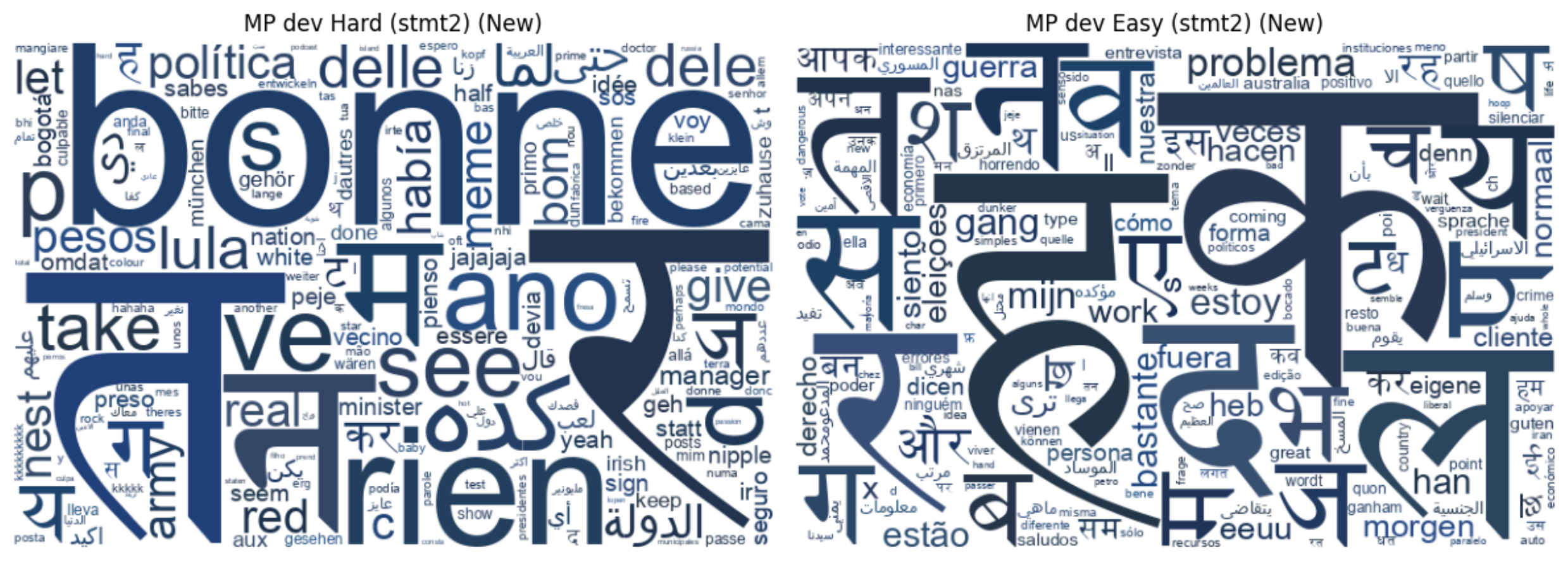}
        \caption{MP (New)}
        \label{fig:wordcloud_mp}
    \end{subfigure}
    \begin{subfigure}{0.9\linewidth}
        \centering
        \includegraphics[width=\textwidth]{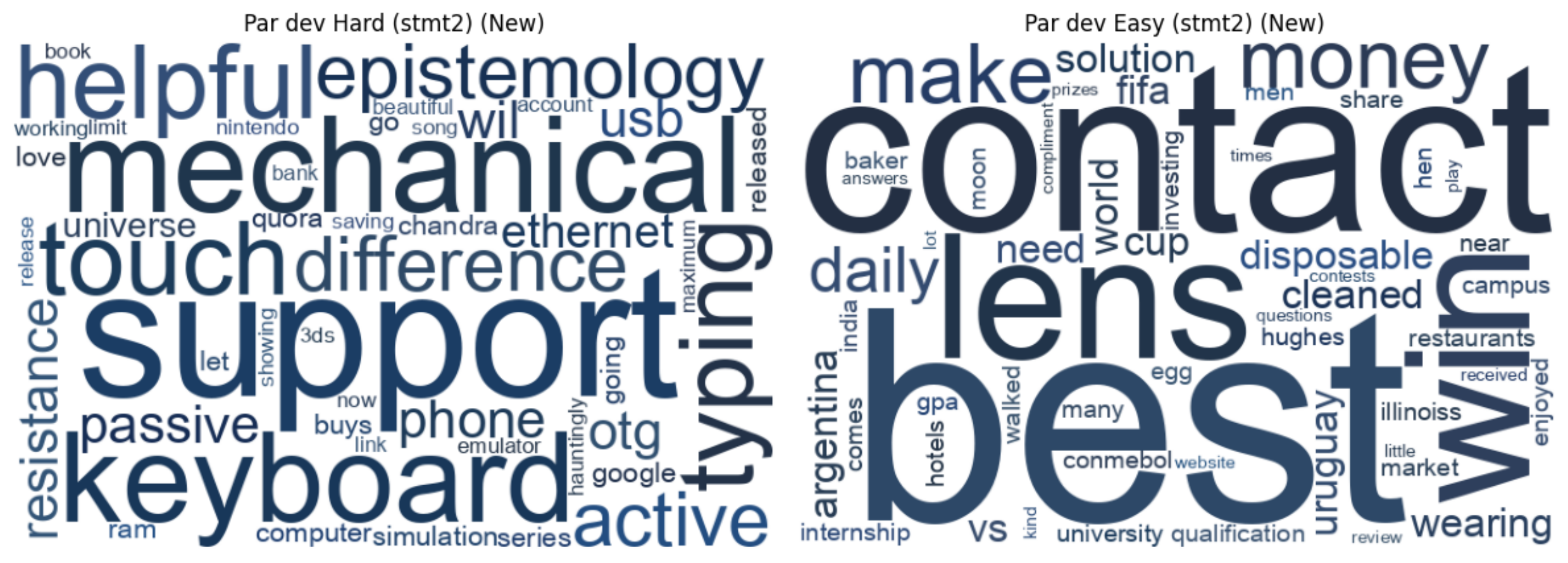}
        \caption{Par (New)}
        \label{fig:wordcloud_par}
    \end{subfigure}
    \caption{Word clouds.}
    \label{fig:wordcloud}
\end{figure}

Word clouds (from the top 25\% hardest and easiest examples (by error) (Figure~\ref{fig:wordcloud}) in each dataset provided further interpretability. In CSC, hard examples in the new system reflected more nuanced social situations (e.g., \textit{“borrowed,” “paid,” “trust”}), while easy examples featured clear sentiment or tonal markers (e.g., \textit{“congrats,” “hang,” “job”}). The new system appeared to better distinguish pragmatic cues of sarcasm. In MP, multilingual word clouds remained dense and difficult to interpret visually, but no major shifts were observed in the most frequent hard/easy terms. Par's clouds showed consistent emphasis on mechanical or structured terms (e.g., \textit{“support,” “contact”}) in hard cases and evaluative language in easy ones (e.g., \textit{“best,” “make,” “win”}). These patterns support the conclusion that the new system is sensitive to social and tonal variation, particularly in CSC.

\subsubsection{Error vs. Token Length and Entropy:}
\begin{figure}[!htb]
    \centering
    \begin{subfigure}{0.9\linewidth}
        \centering
        \includegraphics[width=\textwidth]{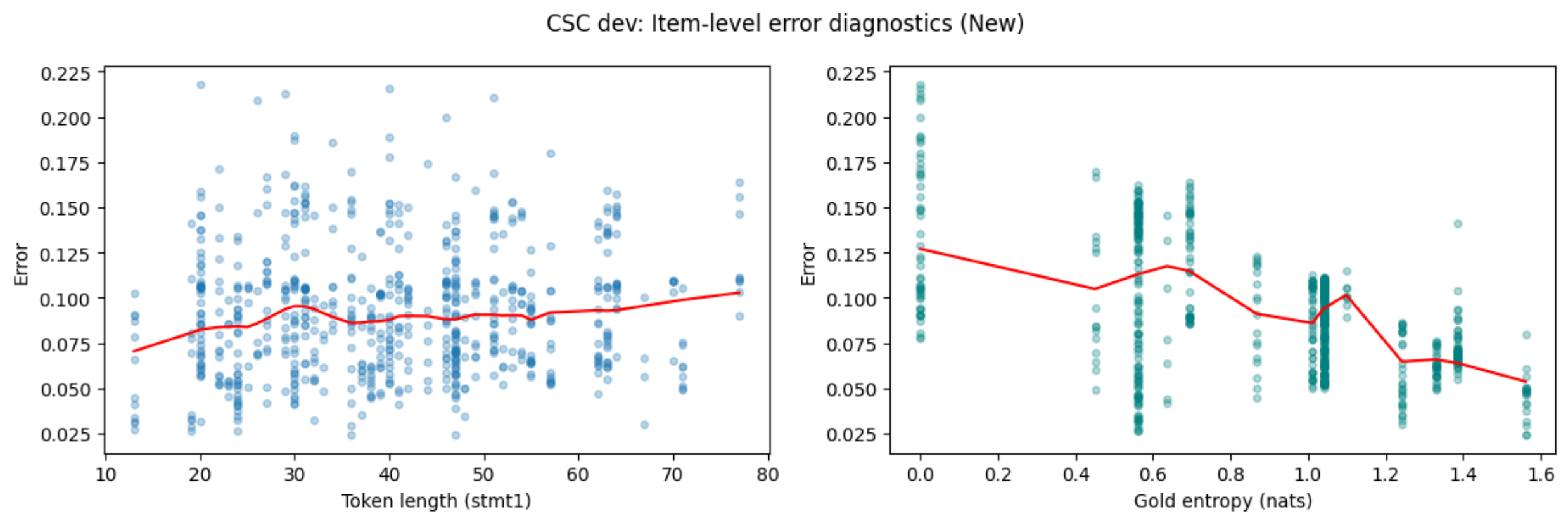}
        \caption{CSC (New)}
    \end{subfigure}
    \begin{subfigure}{0.9\linewidth}
        \centering
        \includegraphics[width=\textwidth]{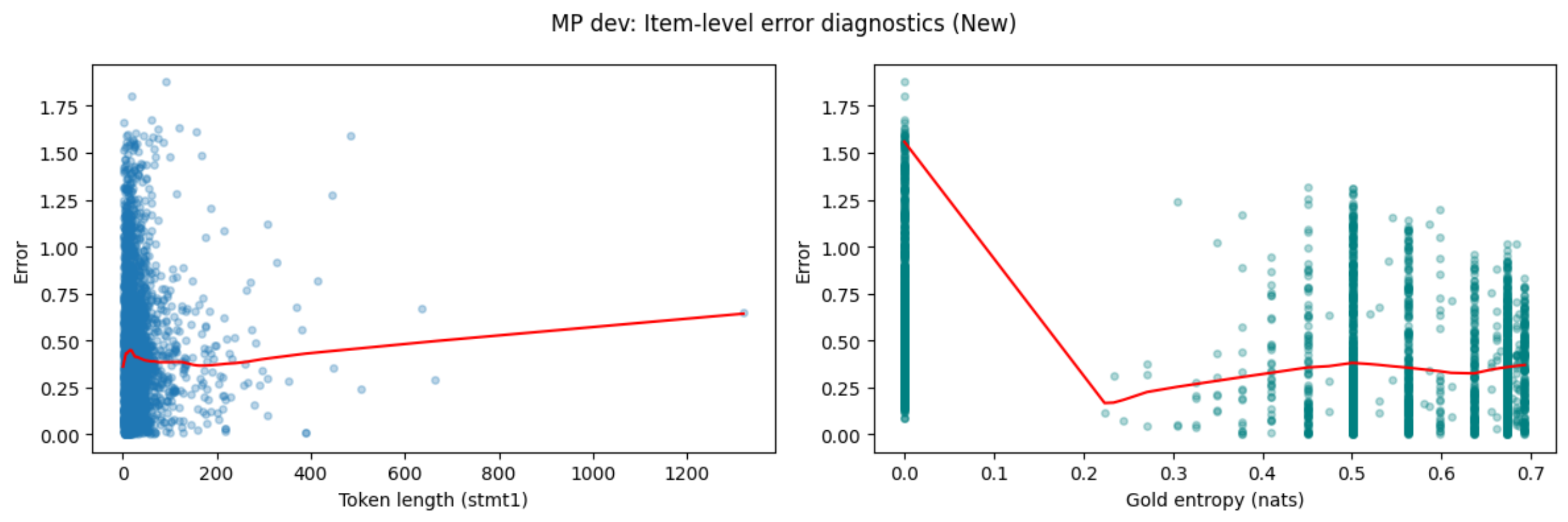}
        \caption{MP (New)}
    \end{subfigure}
    \begin{subfigure}{0.9\linewidth}
        \centering
        \includegraphics[width=\textwidth]{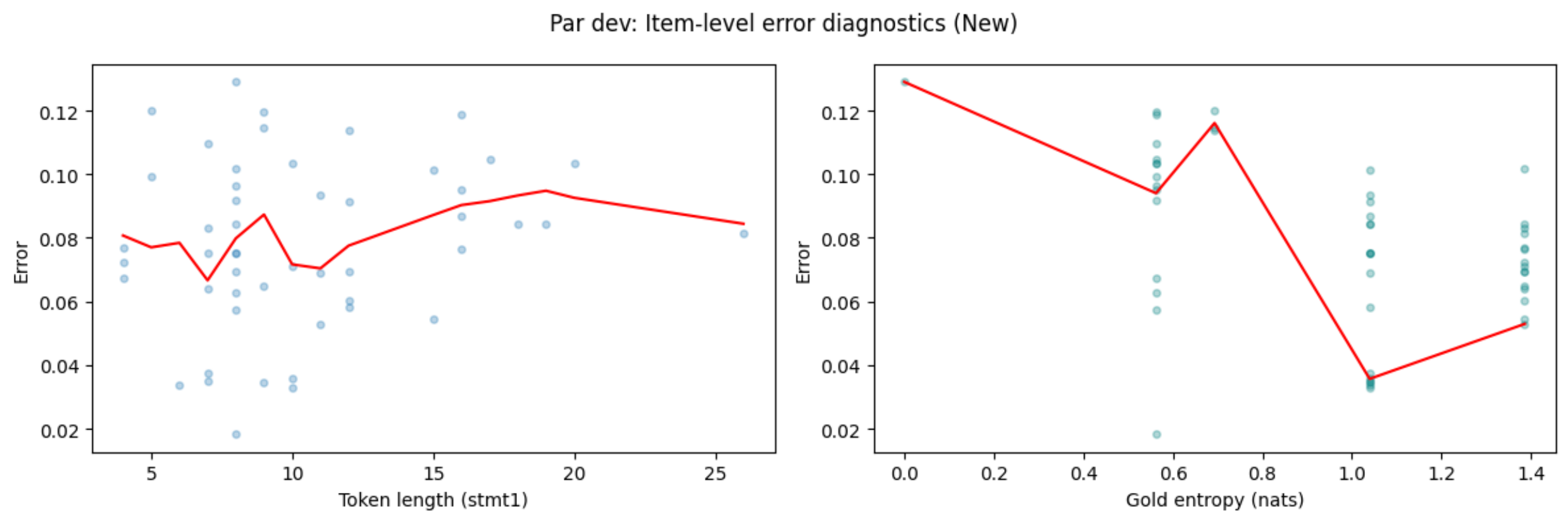}
        \caption{Par (New)}
    \end{subfigure}
    \caption{Error vs. token length and gold entropy across datasets.}
    \label{fig:error_diagnostics}
\end{figure}

Across datasets, we examined how item-level error varied with input length and gold label entropy, refer Figure ~\ref{fig:error_diagnostics}. In CSC, the updated model showed improved behavior on high-entropy items—error steadily decreased as label entropy increased, whereas the original model incurred the highest errors for ambiguous cases. This suggests that the revised model better approximates human uncertainty. A similar trend was observed in MP, although gains were more moderate. For Par, error increased slightly with entropy in the new model, possibly reflecting persistent overfitting to majority-label patterns. Overall, the improved system is more robust to uncertainty in CSC and MP, a key desideratum in perspectivist modeling.

\subsection{Reproducibility - Metadata Prompts}
\label{sec:rep_metadata_prompts}

For full transparency, we provide the exact templates used to verbalize annotator metadata into natural language prompts. These were applied consistently across datasets to ensure reproducible results.

    \paragraph{Par:} The annotator is \texttt{gender}, \texttt{age} years old, from \texttt{nationality} with an education level of \texttt{education}.  
    \paragraph{MP:} The annotator is a \texttt{gender}, \texttt{age} years old, of nationality \texttt{nationality}, born in \texttt{country\_birth} and residing in \texttt{country\_of\_residence}, with student status \texttt{student\_status} and employment status \texttt{employment\_status}, and of \texttt{ethnicity} ethnicity.  
    \paragraph{CSC:} The annotator is a \texttt{gender} and \texttt{age} years old.

These templates allow consistent regeneration of metadata embeddings and support faithful reproduction of our experiments.

\end{document}